\documentclass[runningheads]{llncs}

\usepackage[T1]{fontenc}

\usepackage{hyperref}
\hypersetup{
    colorlinks=true,
    linkcolor=blue,
    filecolor=magenta,      
    urlcolor=cyan
    }
\usepackage{graphicx}
\usepackage{algorithm}
\usepackage{algorithmicx}
\usepackage[noend]{algpseudocode}
\usepackage{booktabs}
\usepackage{amsmath}
\usepackage{amsfonts}
\usepackage{amssymb}
\usepackage{multirow}
\usepackage{xcolor}
\usepackage{pifont}
\usepackage{listings}
\usepackage{varwidth}
\usepackage{float}
\usepackage{adjustbox}

\definecolor{gray}{rgb}{0.4,0.4,0.4}
\definecolor{darkblue}{rgb}{0.0,0.0,0.6}
\definecolor{cyan}{rgb}{0.0,0.6,0.6}
\definecolor{maroon}{rgb}{0.5,0,0}
\definecolor{darkgreen}{rgb}{0,0.7,0}
\definecolor{darkred}{rgb}{0.7,0,0}
\lstdefinestyle{customxml}{
  language=XML,
  basicstyle=\scriptsize\ttfamily,
  columns=fullflexible,
  showstringspaces=false,
  commentstyle=\color{gray}\upshape,
  morestring=[s]{"}{"},
  morecomment=[s]{?}{?},
  morecomment=[s]{!--}{--},
  commentstyle=\color{darkgreen},
  moredelim=[s][\color{black}]{>}{<},
  moredelim=[s][\color{orange}]{\ }{=},
  stringstyle=\color{cyan},
  identifierstyle=\color{cyan}
}

\lstdefinestyle{pythonstyle}{
    language=Python,
    basicstyle=\ttfamily\small,
    keywordstyle=\color{blue}\bfseries,
    commentstyle=\color{green!40!black},
    stringstyle=\color{orange},
    showstringspaces=false,
    tabsize=4,
    backgroundcolor=\color{gray!3},
    captionpos=b,
    breakatwhitespace=false,
    breaklines=true,
    captionpos=b,
    keepspaces=true,
    showspaces=false,
    showstringspaces=false,
    showtabs=false,
    xleftmargin=15pt,
    xrightmargin=15pt,
    numbers=none, 
    frame=none,   
}

\begin{document}
\title{Treat Different Negatives Differently: Enriching Loss Functions with Domain and Range Constraints for Link Prediction}

\titlerunning{Treat Different Negatives Differently}
%
\author{
Nicolas Hubert\inst{1,2}\orcidID{0000-0002-4682-422X} \and
Pierre Monnin\inst{3}\orcidID{0000-0002-2017-8426} \and
Armelle Brun\inst{2}\orcidID{0000-0002-9876-6906} 
\and
Davy Monticolo\inst{1}\orcidID{0000-0002-4244-684X}
}

\authorrunning{N. Hubert et al.}

\institute{Université de Lorraine, ERPI, Nancy, France \and
Université de Lorraine, CNRS, LORIA, Nancy, France \and
Université Côte d’Azur, Inria, CNRS, I3S, Sophia-Antipolis, France
\email{\{nicolas.hubert,armelle.brun,davy.monticolo\}@univ-lorraine.fr}
\email{pierre.monnin@inria.fr}}
%

%
%
\maketitle              
\begin{abstract}
Knowledge graph embedding models (KGEMs) are used for various tasks related to knowledge graphs (KGs), including link prediction. 
They are trained with loss functions that consider batches of true and false triples. 
However, different kinds of false triples exist and recent works suggest that they should not be valued equally, leading to specific negative sampling procedures. 
In line with this recent assumption, we posit that negative triples that are semantically valid w.r.t. signatures of relations (domain and range) are high-quality negatives. 
Hence, we enrich the three main loss functions for link prediction such that all kinds of negatives are sampled but treated differently based on their semantic validity.
In an extensive and controlled experimental setting, we show that the proposed loss functions systematically provide satisfying results 
which demonstrates both the generality and superiority of our proposed approach. 
In fact, the proposed loss functions (1) lead to better MRR and Hits@$10$ values, and (2) drive KGEMs towards better semantic correctness as measured by the Sem@$K$ metric. 
This highlights that relation signatures globally improve KGEMs, and thus should be incorporated into loss functions. Domains and ranges of relations being largely available in schema-defined KGs, this makes our approach both beneficial and widely usable in practice.

\end{abstract}

\section{Introduction}\label{introduction}
A knowledge graph (KG) is a collection of triples $(h,r,t)$ where $h$ (head) and $t$ (tail) are two entities of the graph, and $r$ is a predicate that qualifies the nature of the relation holding between them.
In this work, we do not consider literals.
KGs are inherently incomplete, incorrect, or overlapping and thus major refinement tasks include entity matching and link prediction~\cite{wang2017}. The latter is the focus of this paper.
Link prediction (LP) aims at completing KGs by leveraging existing facts to infer missing ones.
In the LP task, one is provided with a set of incomplete triples, where the missing head (resp. tail) needs to be predicted. This amounts to holding a set of triples where, for each triple, either the head $h$ or the tail $t$ is missing. 

The LP task is often addressed using knowledge graph embedding models (KGEMs).
They represent entities and relations as low-dimensional vectors in a latent embedding space that preserves as much as possible graph structural information and graph properties~\cite{ji2021}.
A plethora of KGEMs has been proposed in the literature. 
They usually differ w.r.t. their scoring function, \textit{i.e.} how they model interactions between entities. 
Entities and relations embeddings are learned throughout several epochs, in an optimization process relying on loss functions. 
Loss functions aims at maximizing the score output by KGEMs for \emph{positive} triples, \textit{i.e.}, triples that exist in the graph, and minimizing the score of \emph{negative} triples, \textit{i.e.}, triples that are absent from the graph.
A recent segment of the literature started investigating the influence of negative triples in the training of KGEMs~\cite{kotnis}.
Indeed, their generation -- called negative sampling -- is usually performed by corrupting true triples, \textit{i.e.}, replacing their head or tail with another entity randomly chosen in the graph.
Enhanced sampling mechanisms were then envisioned and, for example, involve selecting entities of the same type as the entity to replace~\cite{kotnis}.

The latter proposal comes within the scope of works considering the underpinning semantics of KGs as additional information to improve results w.r.t. the LP task. 
Over the past few years, semantic information has been incorporated in various parts of KGEMs, \textit{e.g.}, as mentioned, the negative sampling procedure~\cite{jain_iswc,krompas}, but also the model itself~\cite{tarp,transc,autoeter,transet,tkrl}, or the loss function~\cite{cao2022,damato2021,guo2015,minervini2017}. 
Existing works proposing to include semantic information into loss functions showcase promising results.
However, they are restricted to specific loss functions~\cite{damato2021,guo2015}, or consider ontological axioms~\cite{damato2021,minervini2017} that do not include domain and range axioms (or relation signatures) which are widely available in KGs.
Interestingly, such axioms could be leveraged to generate different kinds of negative triples, \textit{i.e.}, negative triples that respect relation signatures (called semantically valid) and those that do not (called semantically invalid).
Additionally, to the best of our knowledge, negative triples were only studied in the sampling procedure, where specific ones are chosen to train on and the others are discarded. 
The possibility to sample all kinds of negatives but differently consider them within loss functions was left unassessed.
This twofold observation motivates our first research question:
\begin{description}
    \item[RQ1] how main loss functions used in LP can incorporate
    domain and range constraints to differently consider negative triples based on their semantic validity? 
\end{description}

Precedent work also pointed out the performance gain of incorporating ontological information as measured by rank-based metrics such as MRR and Hits@$K$~\cite{cao2022,damato2021,guo2015,minervini2017}. 
However, while such approaches include semantic information as KGEM inputs, the semantic capabilities of the resulting KGEM are left unassessed, even though this would provide a fuller picture of its performance~\cite{dl4kg,hubert2023}. Hence, our second research question:
\begin{description}
    \item[RQ2] what is the impact of incorporating relation signatures into loss functions on the overall KGEM performance?
\end{description}

To address both questions, we propose signature-driven loss functions, \textit{i.e.} loss functions containing terms that depend on some background knowledge (BK) about types of entities and domains and ranges of relations. 
To broaden the impact of our approach, our work is concerned with the three most encountered loss functions in the literature: the pairwise hinge loss (PHL)~\cite{transe}, the 1-N binary cross-entropy loss (BCEL)~\cite{conve}, and the pointwise logistic loss (PLL)~\cite{complex} (further detailed in Section~\ref{lossfunc}).
For each of them, we propose a tailored signature-driven version.
The considered BK is available in many schema-defined KGs~\cite{ding2018}, which makes these newly introduced loss functions widely usable in practice.
 Furthermore, the impact of loss functions is evaluated using both rank-based metrics and Sem@$K$~\cite{dl4kg,hubert2023} -- a metric that measures the consistency of KGEMs predictions for the LP task with relation signatures, \textit{i.e.}, the semantic correctness of predictions. 

To summarize, the main contributions of this work are: 
\begin{itemize}
    \item We propose signature-driven versions for the three mostly used loss functions for the LP task, leveraging BK about relation domains and ranges.
    \item We evaluate our approach in terms of traditional rank-based metrics, and also w.r.t. Sem@$K$, which gives more insight into the benefits of our proposal.
    \item We show that the designed signature-driven loss functions provide, in most cases, better performance w.r.t. both rank-based metrics and Sem@$K$. Consequently, our findings strongly indicate that signature information should be systematically incorporated into loss functions.
\end{itemize}
The remainder of the paper is structured as follows. 
Related work is presented in Section~\ref{related-work}. 
In Section~\ref{lossfunc}, we detail the signature-driven loss functions proposed in this work. 
Dataset descriptions and experimental settings are provided in Section~\ref{experimental-setting}. Key findings are presented in Section~\ref{results} and are further discussed in Section~\ref{discussion}.
Lastly, Section~\ref{conclusion} sums up the main findings and outlines future research directions.

\section{Related Work}\label{related-work}
This section firstly relates to former contributions that make use of semantic information to enhance model results regarding the LP task. Emphasis is placed on how semantic information can be incorporated in the loss functions~(Section~\ref{related-work:semantic}) -- a research avenue which remains relatively unexplored compared to incorporating semantic information in other parts of the learning process, \textit{e.g.} in the negative sampling or in the interaction function. Secondly, a brief background on the mainstream loss functions is provided~(Section~\ref{related-work:lossfunc}). This is to help position our contributions w.r.t. the vanilla loss functions used in practice.
\subsection{Semantic-Enhanced Approaches}\label{related-work:semantic}
A significant body of the literature proposes approaches that incorporate semantic information for performing LP with KGEMs, with the purpose of improving KGEM performance w.r.t. traditional rank-based metrics. 

The most straightforward way to do so is to embed semantic information in the model itself. For instance, AutoETER~\cite{autoeter} is an automated type representation learning mechanism that can be used with any KGEM and that learns the latent type embedding of each entity. In TaRP~\cite{tarp}, type information and instance-level information are simultaneously considered and encoded as prior probabilities and likelihoods of relations, respectively.
TKRL~\cite{tkrl} bridges type information with hierarchical information: while type information is utilized as relation-specific constraints, hierarchical types are encoded as projection matrices for entities. TKRL allows entities to have different representations in different types. Similarly, in~\cite{transet}, the proposed KGEM allows entities to have different vector representations depending on their respective types. TransC~\cite{transc} encodes each concept of a KG as a sphere and each instance as a vector in the same semantic space. The relations between concepts and instances (\texttt{rdf:type}), and the relations between concepts and sub-concepts (\texttt{rdfs:subClassOf}) are based on the relative distance within this shared semantic space.

A few works incorporate semantic information to constrain the negative sampling (NS) procedure and generate meaningful negative triples~\cite{krompas,jain_iswc,weyns}. For instance, type-constrained negative sampling (TCNS)~\cite{krompas} replaces the head or the tail of a triple with a random entity belonging to the same type (\texttt{rdf:type}) as the ground-truth entity. Jain \textit{et al.}~\cite{jain_iswc} go a step further and use ontological reasoning to iteratively improve KGEM performance by retraining the model on inconsistent predictions.
It is noteworthy that our approach adopts an orthogonal direction of such works by proposing to sample all kinds of negatives but to treat them differently in the loss function when training.

A few work actually propose to include semantic information in the learning and optimization process.
In~\cite{guo2015}, entities embeddings of the same semantic category are enforced to lie in a close neighborhood of the embedding space. However, their approach only fits single-type KGs. In addition, the only mainstream model benchmarked in this work is TransE~\cite{transe}, and only the pairwise hinge loss is used. Likewise, d'Amato \textit{et al.}~\cite{damato2021} solely consider the pairwise hinge loss, and their approach is benchmarked w.r.t. to translational models only. Moreover, BK is injected in the form of \texttt{equivalentClass}, \texttt{equivalentProperty}, \texttt{inverseOf}, and \texttt{subClassOf} axioms, similarly to~\cite{minervini2017} who incorporate \texttt{equivalentProperty} and \texttt{inverseOf} axioms as regularization terms in the loss function. However, the aforementioned axioms are rarely provided in KGs~\cite{ding2018}. Cao \textit{et al.}~\cite{cao2022} propose a new regularizer called Equivariance Regularizer, which limits overfitting by using semantic information. However, their approach is data-driven and does not rely on a schema. In contrast, the approach presented in Section~\ref{lossfunc} leverages domain and range constraints which are available in most schema-defined KGs.

Finally, all the aforementioned semantic-driven approaches are only evaluated w.r.t.. rank-based metrics. However, semantic-driven approaches would benefit from a semantic-oriented evaluation. To the best of our knowledge, the work around Sem@$K$~\cite{dl4kg,ekaw,hubert2023} is the only one to provide appropriate tools for measuring KGEM semantic correctness. Hence, our experiments will also be evaluated with this metric.

\subsection{Loss Functions for the Link Prediction Task}\label{related-work:lossfunc}
Few works revolve around the influence of loss functions on KGEM performance~\cite{ali2022,mohamed2019,mohamed2021}. Mohamed~\textit{et al.}~\cite{mohamed2019} point out the lack of consideration regarding the impact of loss functions on KGEM performance.
Experimental results provided in~\cite{ali2022} indicate that no loss function consistently provides the best results, and that it is rather the combination between the scoring and loss functions that impacts KGEM performance. In particular, some scoring functions better match with specific loss functions. For instance, Ali~\textit{et al.} show that TransE can outperform state-of-the-art KGEMs when configured with an appropriate loss function. Likewise, Mohamed~\textit{et al.}~\cite{mohamed2019} show that the choice of the loss function significantly influence KGEM performance. Consequently, they provide an extensive benchmark study of the main loss functions used in the literature. Namely, their analysis relies on a commonly accepted categorization between pointwise and pairwise loss functions. 
The main difference between pointwise and pairwise loss functions lies in the way the scoring function, the triples, and their respective labels are considered all together. Under the pointwise approach, the loss function relies on the predicted scores for triples and their actual label values, which is usually $1$ for positive triple and $0$ (or $-1$) for negative triples. In contrast, pairwise loss functions are defined in terms of differences between the predicted score of a true triple and the score of a negative counterpart. 

In our approach (Section~\ref{lossfunc}), we consider the three most commonly used loss functions for performing LP~\cite{rossi}: the pairwise hinge loss (PHL)~\cite{transe}, the 1-N binary cross-entropy loss (BCEL)~\cite{conve}, and the pointwise logistic loss (PLL)~\cite{complex}. Their vanilla formulas are recalled in Equations~\eqref{eq:phl-vanilla}, \eqref{eq:bcel-vanilla}, and \eqref{eq:pll-vanilla}.

\begin{equation}
\mathcal{L}_{PHL} = \sum_{t\in\mathcal{T^{+}}} \sum_{t'\in\mathcal{T^{-}}}  
\left[ \gamma + f(t') - f(t) \right]_{+}
\label{eq:phl-vanilla}
\end{equation}
where $\mathcal{T}$, $f$, and $[x]_{+}$ denote a batch of triples, the scoring function, and the positive part of $x$, respectively. $\mathcal{T}$ is further split into a batch of positive triples $\mathcal{T^{+}}$ and a batch of negative triples $\mathcal{T^{-}}$. $\gamma$ is a configurable margin hyperparameter specifying how much the scores of positive triples should be separated from the scores of corresponding negative triples.

\begin{equation}
\mathcal{L}_{BCEL} = -\frac{1}{|\mathcal{E}|}\sum_{t\in\mathcal{T}}
\ell(t)\log(f(t)) + (1-\ell(t))\log(1-f(t))
\label{eq:bcel-vanilla}
\end{equation}
where $\ell(t) \in \{0,1\}$ denotes the true label of $t$ and $\mathcal{T}$ is a batch with all possible $(h,r,*)$. $|\mathcal{E}|$ is the number of entities in the KG.

\begin{equation}
\mathcal{L}_{PLL} = \sum_{t\in\mathcal{T}}
\log(1+\exp^{-\ell(t) \cdot f(t)})
\label{eq:pll-vanilla}
\end{equation}
where $\ell(t) \in \{-1,1\}$ denotes the true label of $t$.

\section{Signature-driven Loss Functions}\label{lossfunc}
Building on the limits of previous work (Section~\ref{related-work:semantic}), we propose signature-driven loss functions that extend the three most frequently used loss functions~\cite{rossi} and leverage BK about domains and ranges of relations, which are provided in many KGs used in the literature.

The purpose of the proposed loss functions is to distinguish \emph{semantically valid negatives} from \emph{semantically invalid ones}. The former are defined as triples $(h,r,t)$ respecting both the domain and range of the relation $r$, \textit{i.e.}, $$\operatorname{type}(h) \cap \operatorname{domain}(r) \neq \emptyset \land \operatorname{type}(t) \cap \operatorname{range}(r) \neq \emptyset$$ whereas the latter violate at least one of the constraints, \textit{i.e.} $$\operatorname{type}(h) \cap \operatorname{domain}(r) = \emptyset \lor \operatorname{type}(t) \cap \operatorname{range}(r) = \emptyset.$$ $\operatorname{domain}(r)$ (resp. $\operatorname{range}(r)$) is defined as the expected type as head (resp. tail) for the relation $r$. For example, the relation \texttt{presidentOf} expects a \texttt{Person} as head and a \texttt{Country} as tail. Starting from a positive triple (\texttt{EmmanuelMacron}, \texttt{presidentOf}, \texttt{France}) which represents a true fact, (\texttt{BarackObama}, \texttt{presidentOf}, \texttt{France}) and (\texttt{EmmanuelMacron}, \texttt{presidentOf}, \texttt{Germany}) are examples of semantically valid negative triples, whereas (\texttt{Adidas}, \texttt{presidentOf}, \texttt{France}) and (\texttt{EmmanuelMacron}, \texttt{presidentOf}, \texttt{Christmas}) are examples of semantically invalid negative triples. In this work, entities are multi-typed. Therefore, $\operatorname{type}(e)$ returns the set of types (a.k.a. classes) that the entity $e$ belongs to.

We introduce a loss-independent $\epsilon$ factor, which is dubbed as the \textit{semantic factor} and aims at bringing the scores of semantically valid negative triples closer to the positive ones. This common semantic factor fitting into different loss functions shows the generality of our approach that can possibly be extended to other loss functions.
Interestingly, this factor also allows to take into account to some extent the Open World Assumption (OWA) under which KGs are represented. 
Under the OWA, triples that are not represented in a KG are either false or missing positive triples. 
In traditional training procedures, these triples are indiscriminately considered negative, which corresponds to the Closed World Assumption.
On the contrary, our proposal considers semantically invalid triples as true negative while semantically valid triples (and possibly missing positive or false negative under the OWA) are closer to true positive triples.
This assumes that entity types are complete and correct. 
	
$\mathbf{\mathcal{L}_{PHL}}$ defined in Equation~\eqref{eq:phl-vanilla} relies on the margin hyperparameter $\gamma$. Increasing (resp. descreasing) the value of $\gamma$ will increase (resp. descrease) the margin that will be set between the scores of positive and negative triples.
However, this unique $\gamma$ treats all negative triples indifferently: the same margin will separate the scores of semantically valid and semantically invalid negative triples from the score of the positive triple they both originate from. 
We suggest that the scores of these two kinds of negative triples should be treated differently.
Hence, our approach redefines $\mathcal{L}_{PHL}$ as follows (Equation~\eqref{eq:phl-sem}):
\begin{equation}
\begin{split}
\mathcal{L}^{S}_{PHL} & = \sum_{t\in\mathcal{T^{+}}} \sum_{t'\in\mathcal{T^{-}}}
\left[ \gamma \cdot \ell(t') + f(t') - f(t) \right]_{+}  \\ \\
\text{where } & \ell(t') = \begin{cases}
    1 \text{ if $t'$ is semantically invalid} \\
    \epsilon \text{ otherwise}
\end{cases}
\label{eq:phl-sem}
\end{split}
\end{equation}
The loss function in Equation~\eqref{eq:phl-sem} now has a superscripted \textit{S} to make it clear this is the signature-driven version of the vanilla $\mathcal{L}_{PHL}$ as defined in Equation~\eqref{eq:phl-vanilla}. 
A choice of $\epsilon < 1$ leads the KGEM to apply a higher margin between scores of positive and semantically invalid triples than between positive and semantically valid ones. 
For a given positive triple, this allows to keep the scores of its semantically valid negative counterparts relatively closer compared to the scores of its semantically invalid counterparts. Intuitively, when the KGEM outputs wrong predictions, more of them are still expected to meet the domain and range constraints imposed by relations. Thus, wrong predictions are assumed to be more meaningful, and, in a sense, semantically closer to the ground-truth triple. 
	
$\mathbf{\mathcal{L}_{BCEL}}$ defined in Equation~\eqref{eq:bcel-vanilla} is adapted to $\mathcal{L}^{S}_{BCEL}$ by redefining the labelling function $\ell$. In particular, when dealing with a KG featuring typed entities and providing information about domains and ranges of relations, the labelling function $\ell$ is no longer binary. Instead, the labels of semantically valid negative triples can be fixed to some intermediate value between the label value of positive triples and of semantically invalid negative triples, which leads to the labelling function $\ell$ defined in Equation~\eqref{eq:label-bcel}:
\begin{equation}
\label{eq:label-bcel}
\begin{aligned}
\ell(t') = \left\{
    \begin{array}{lll}
1 & \mbox{if } t' \in  \mathcal{T^{+}}\\
 \epsilon & \mbox{if } t' \in \mathcal{T^{-}} \text{ and $t'$ is semantically valid} \\ 
 0 & \mbox{if } t' \in \mathcal{T^{-}} \text{ and $t'$ is semantically invalid}
    \end{array}
\right.
\end{aligned}
\end{equation}
\noindent where the semantic factor $\epsilon$ is a tunable hyperparameter denoting the label value of semantically valid negative triples. The intuition underlying the refinement of the labelling function $\ell$ is to voluntarily cause some confusion between semantically valid negative triples and positive triples.
By bridging their respective label values, it is expected that the KGEM will somehow consider the former as ``less negative triples`` and assign them a higher score compared to positive triples.

$\mathbf{\mathcal{L}_{PLL}}$ defined in Equation~\eqref{eq:pll-vanilla} could be adapted to $\mathcal{L}^{S}_{PLL}$ similarly to $\mathcal{L}^{S}_{BCEL}$. In other words, the labelling function $\ell$ could also output an intermediate label value $\epsilon$ for semantically valid negative triples. Although this approach provides very good results in terms of Sem@$K$ values, it does not provide consistently good results across datasets.
Furthermore, obtained results in terms of MRR and Hits@$K$ can be far below the ones obtained with the vanilla model (see supplementary materials for further details).
That is why, here, to treat semantically valid and invalid negative triples differently, instead of modifying the labelling function $\ell$, the semantic factor $\epsilon$ for $\mathcal{L}^{S}_{PLL}$ defines the probability with which semantically valid negative triples are considered as positive triples and therefore are labelled the same way. For example, with $\epsilon = 0.05$, at each training epoch and for each batch, the semantically valid negative triples of the given training batch will be considered positive with a probability of $5\%$.

It is noteworthy that our approach can be applied in practice to KGs with or without types, domains and ranges.
Indeed, in the absence of such background knowledge, our signature-driven loss functions reduce to their respective vanilla counterparts.
Recall that our approach does not focus on negative sampling or complex negative sample generators such as KBGAN~\cite{kbgan}, NSCaching~\cite{nscaching}, and self-adversarial negative sampling~\cite{rotate}. Although these works are related to ours, they constrain negative sampling upstream.
On the contrary, our approach does not constrain the sampling of negative triples but rather dynamically distributes the negative triples into different parts of the loss functions, based on their semantic validity.

\section{Experimental Setting}\label{experimental-setting}

\subsection{Evaluation Metrics}
In our experiments, KGEM performance is assessed w.r.t. MRR, Hits@$K$ and Sem@$K$, with $K=10$.
\textbf{Mean Reciprocal Rank (MRR)} corresponds to the arithmetic mean over the reciprocals of ranks of the ground-truth triples.
MRR is bounded in the $\left[0,1\right]$ interval, where the higher the better.
\textbf{Hits@$K$} accounts for the proportion of ground-truth triples appearing in the first $K$ top-scored triples.
This metric is bounded in the $ \left[0,1\right]$ interval and its values increases with $K$, where the higher the better.
\textbf{Sem@$K$}\cite{dl4kg,hubert2023} accounts for the proportion of
triples that are semantically valid in the first $K$ top-scored triples:
\begin{equation}
\mathrm{Sem} @ {K}=\frac{1}{|\mathcal{B}|} \sum_{q \in \mathcal{B}}\frac{1}{{K}}
\sum_{q' \in \mathcal{S}^{K}_q} \operatorname{compatibility}(q,q')
\label{eq:sematk}
\end{equation} 
where, given a ground-truth triple $q = (h,r,t)$, $\mathcal{S}^{K}_q$ is the list of the top-$K$ candidate triples scored by a KGEM (\textit{i.e.}, by predicting the tail for $(h,r,?)$ or the head for $(?,r,t)$). 
A candidate triple $q'$ is assessed 
w.r.t. $q$ by the $\operatorname{compatibility}$ function
that checks whether the predicted head (resp. tail) belongs to the domain (resp. range) of the relation. 
In this work, class hierarchies are considered: if a relation has a given class as domain (resp. range), entities from its subclasses are considered semantically valid.
Sem@$K$ is bounded in the $\left[0,1\right]$ interval. 

\subsection{Datasets and Models}\label{datasets}

Even though our approach could be applied in KGs with or without relation signatures, we evaluate our proposal in an ideal experimental setting to precisely qualify the interest of considering domains and ranges.
Firstly, all relations appearing in the training set have a defined domain and range. Secondly, both the head $h$ and tail $t$ of train triples have another semantically valid counterpart for negative sampling. These two conditions guarantee that each positive train triple can be paired with at least one semantically valid triple. Finally, validation and test sets contain triples whose relation have a well-defined domain (resp. range), as well as more than 10 semantically valid candidates as head (resp. tail). This ensures Sem@$K$ is not unduly penalized and can be calculated on the same set of entities as Hits@$K$ and MRR until $K=10$.

To ensure these requirements, we filtered FB15k237-ET, DBpedia93k, and YAGO4-19k~\cite{hubert2023} so that they comply with the aforementioned criteria.
In the following, their filtered versions are referred to as FB15k187, DBpedia77k, and Yago14k, respectively. 
Table~\ref{tab:datasets} provides statistics for these datasets and reflects the diversity of their characteristics. In this work, several KGEMs are considered:
TransE~\cite{transe}, TransH~\cite{transh}, and DistMult~\cite{distmult} using $\mathcal{L}_{PHL}$; ComplEx~\cite{complex} and SimplE~\cite{simple} using $\mathcal{L}_{PLL}$;  ConvE~\cite{conve}, TuckER~\cite{tucker}, and RGCN~\cite{rgcn} using $\mathcal{L}_{BCEL}$.
Datasets and codes are available in our GitHub repository.\footnote{\label{git}\url{https://github.com/nicolas-hbt/semantic-lossfunc/}}

\begin{table}
\centering
\caption{Datasets used in the experiments. These are filtered versions of the standard FB15k237-ET, DBpedia93k, and YAGO4-19k.}\label{tab:datasets}
\begin{tabular}{lrrrrrr@{/}r@{/}r}
\toprule
Dataset & \multicolumn{1}{c}{$|\mathcal{E}|$} & \multicolumn{1}{c}{$|\mathcal{R}|$} & \multicolumn{1}{c}{$|\mathcal{T}_{train}|$} & \multicolumn{1}{c}{$|\mathcal{T}_{valid}|$} & \multicolumn{1}{c}{$|\mathcal{T}_{test}|$} & \multicolumn{3}{c}{Split ratios}\\
\midrule
FB15k187 & $14,305$ & $187$ & $245,350$ & $15,256$ & $17,830$ & $88$\% & $5.5$ \% &$6.5$\%\\
DBpedia77k  & $76,651$ & $150$ & $140,760$ & $16,334$ & $32,934$ & $74$\% & $9$ \% & $17$\%\\
Yago14k & $14,178$ & $37$ & $18,263$ & $472$ & $448$ & $95$\% & $2.5$\% & $2.5$\%  \\
\bottomrule
\end{tabular}
\end{table}

\subsection{Implementation Details}\label{implementation}
For the sake of comparison, MRR, Hits@$K$ and Sem@$K$ are computed after training models from scratch.
KGEMs used in the experiments were implemented in PyTorch. After training KGEMs for a large number of epochs, we noticed the best achieved results were found around epoch $400$ or below. Consequently, a maximum of $400$ epochs of training was set, as in LibKGE\footnote{\url{https://github.com/uma-pi1/kge/}} (except RGCN which is trained during 4,000 epochs due to lower convergence to the best achieved results).
Except when using the BCEL which does not require negative sampling, Uniform Random Negative Sampling~\cite{transe} was used to pair each train triple with two corresponding negative triples: one which is semantically invalid, and one which is semantically valid. 
Regarding BCEL, each positive triple is scored against negative triples formed with all other entities in the graph. 
Hence, negative triples comprise both semantically valid and invalid triples.
It should be noted that the best epsilon values for each combination of model and dataset were found on the validation sets. Once the best epsilon values are found, they remain fixed for all triples.
In order to ensure fair comparisons between models, embeddings are initialized with the same seed and each model is fed with exactly the same set of negative triples at each epoch.
Grid-search based on predefined hyperparameters was performed. 
Best hyperparameters and the  full hyperparameter space are reported on GitHub.\textsuperscript{\ref{git}}

\section{Results}\label{results}
\subsection{Global Performance}\label{global-perf}
\begin{table}[h]
        \caption{Rank-based and semantic-based results. Bold fonts indicate which model performs best w.r.t. a given metric. Suffixes V and S indicate whether the model is trained under the vanilla or signature-driven version of the loss function, respectively. Hits@$10$ and Sem@$10$ are abbreviated to H@$10$ and S@$10$. Underlined cells indicate results that are more specifically referred to in Section~\ref{global-perf}. We use the symbol $^\dag$ when the comparison of results in a duel (\textit{e.g.} TransE-V vs. TransE-S) is statistically significant ($\alpha = .05$) for a given metric and dataset.}
	\label{tab:schema-defined-results-compiled}
	\centering
                \setlength{\tabcolsep}{0.15cm}
                \begin{adjustbox}{width=0.9\textwidth}
			\begin{tabular}{lccccccccc}
                    \toprule
				&\multicolumn{3}{c}{FB15k187} & \multicolumn{3}{c}{DBpedia77k} & \multicolumn{3}{c}{Yago14k}
                \\
                    & MRR & H@10 & S@10 &
                    MRR & H@10 & S@10 &
                    MRR & H@10 & S@10 \\
				\toprule
				TransE-V &
    $.260$&$.446$&$.842$& $.274$&$.438$&$.936$& $.868$&$\mathbf{.945}$&$.795$\\
                TransE-S &
    $\mathbf{.315}^\dag$&$\mathbf{.497}^\dag$&$\mathbf{.973}^\dag$& $\mathbf{.275}$&$\mathbf{.440}$&$\mathbf{.985}^\dag$& $\mathbf{.876}$&$.944$&$\mathbf{.968}^\dag$ \\
    \midrule
                TransH-V &
    $.266$&$.450$&$.855$& $.270$&$.437$&$.907$& $.836$&$.944$&$.581$ \\
                TransH-S &
    $\mathbf{.319}^\dag$&$\mathbf{.501}^\dag$&$\mathbf{.973}^\dag$& $\mathbf{.274}^\dag$&$\mathbf{.442}^\dag$&$\mathbf{.980}^\dag$& $\mathbf{.857}^\dag$&$\mathbf{.945}$&$\underline{\mathbf{.831}}^\dag$ \\
    \midrule
                DistMult-V &
    $.291$&$.457$&$.824$& $.295$&$.405$&$.784$& $.904$&$\mathbf{.930}$&$.409$ \\
			    DistMult-S &
    $\mathbf{.332}^\dag$&$\mathbf{.504}^\dag$&$\mathbf{.971}^\dag$& $\mathbf{.300}^\dag$&$\mathbf{.416}^\dag$&$\mathbf{.901}^\dag$& $\mathbf{.912}^\dag$&$.929$&$\mathbf{.449}^\dag$ \\
    \midrule
    		ComplEx-V &
    $.280$&$.416$&$.472$& $\mathbf{.309}^\dag$&$\mathbf{.415}^\dag$&$.769$& $\mathbf{.925}$&$\mathbf{.932}$&$.333$ \\
    		ComplEx-S &
    $\mathbf{.316}^\dag$&$\mathbf{.476}^\dag$&$\underline{\mathbf{.796}}^\dag$& $.297$&$.409$&$\mathbf{.897}$& $.923$&$.931$&$\underline{\mathbf{.667}}^\dag$ \\
    \midrule
                SimplE-V &
    $.261$&$.387$&$.462$& $\mathbf{.259}^\dag$&$\mathbf{.346}^\dag$&$\mathbf{.883}^\dag$& $\mathbf{.926}$&$\mathbf{.931}$&$.355$ \\
                SimplE-S &
    $\mathbf{.268}^\dag$&$\mathbf{.409}^\dag$&$\underline{\mathbf{.759}}^\dag$& $.230$&$.302$&$.850$& $.924$&$.927$&$\underline{\mathbf{.769}}^\dag$ \\
    \midrule
				ConvE-V &
    $.273$&$.470$&$.973$& $.273$&$.382$&$.935$& $\mathbf{.934}$&$\mathbf{.942}$&$.904$ \\
				ConvE-S &
    $\mathbf{.283}^\dag$&$\mathbf{.476}^\dag$&$\mathbf{.996}^\dag$& $\mathbf{.283}^\dag$&$\mathbf{.405}^\dag$&$\mathbf{.985}^\dag$& $.933$&$.940$&$\mathbf{.997}^\dag$  \\
    \midrule
    			TuckER-V &
    $.316$&$.516$&$.985$& $.311$&$.410$&$.912$& $.923$&$.927$&$.781$ \\
                TuckER-S &
    $\mathbf{.320}^\dag$&$\mathbf{.522}^\dag$&$\mathbf{.996}^\dag$& $\mathbf{.312}$&$\mathbf{.421}^\dag$&$\mathbf{.969}^\dag$& $\mathbf{.931}^\dag$&$\mathbf{.943}^\dag$&$\mathbf{.929}^\dag$ \\
    \midrule
                RGCN-V &
    $.241$&$.386$&$.775$& $.194$&$.297$&$.872$& $.911$&$.923$&$.349$ \\
                RGCN-S &
    $\mathbf{.260}^\dag$&$\mathbf{.415}^\dag$&$\mathbf{.860}^\dag$& $\mathbf{.197}^\dag$&$\mathbf{.320}^\dag$&$\mathbf{.957}^\dag$& $\mathbf{.927}^\dag$&$\mathbf{.934}^\dag$&$\underline{\mathbf{.828}}^\dag$ \\
    \bottomrule
			\end{tabular}
   \end{adjustbox}
 \end{table}

Table~\ref{tab:schema-defined-results-compiled} displays KGEM performance, datasets and evaluation metrics of interest. We performed t-tests (when prediction-related data follow a normal distribution according to Shapiro test) and Wilcoxon tests (when they do not) at the significance level $\alpha = .05$. Table~\ref{tab:schema-defined-results-compiled} clearly shows that, with the sole exception of SimplE on DBpedia77k, the signature-driven loss functions $\mathcal{L}^{S}_{PHL}$, $\mathcal{L}^{S}_{BCEL}$, and $\mathcal{L}^{S}_{PLL}$ all lead to significant improvement in terms of Sem@$10$. Importantly, in some cases the relative gain in Sem@$10$ compared to the corresponding vanilla loss function is huge (underlined in Table~\ref{tab:schema-defined-results-compiled}): $+137\%$, $+117\%$, and $+100\%$ for RGCN, SimplE, and ComplEx on the smaller Yago14k dataset, respectively, and $+69\%$ and $+64\%$ for ComplEx and SimplE on FB15k187, respectively. These gains in terms of Sem@$10$ are observed regardless of the loss function at hand, which demonstrates that our designed signature-driven loss functions can all drive KGEM semantic correctness towards more satisfying results. It is worth noting that, in most cases, they also lead to better KGEM performance as measured by MRR and Hits@$10$. We observe that in $19$ out of $24$ ($\approx79\%$) one-to-one comparisons between the same KGEM trained with vanilla vs. signature-driven loss functions, better MRR values are reported for the model trained under the signature-driven loss function. 
On the remaining comparisons ($\approx21\%$), only two highlight statistically significant losses in terms of MRR.
However, they are minimal and often for the benefit of significantly better Sem@$10$ values.
This observation raises the question whether better MRR and Hits@$K$ should be pursued at any cost, or whether a small drop w.r.t. to these metrics is acceptable if this leads to a significantly better KGEM semantic correctness (see Section~\ref{section:kgem-eval}). Plus, these promising results imply that even if the intended purpose is to only maximize MRR and Hits@$K$ values, taking the available signature information into consideration is strongly advised: this does not only improve KGEM semantic correctness, but also provide performance gains in terms of MRR and Hits@$K$. 

In the following, we provide a finer-grained results' analysis, which focuses on the different loss functions and datasets used in the experiments.
Although we previously showed the effectiveness of our approach, the benefits brought from considering BK in the form of relation domains and ranges differ across loss functions. In particular, the gains achieved using $\mathcal{L}^{S}_{PHL}$ and $\mathcal{L}^{S}_{BCEL}$ are substantial. With these loss functions, we observe a systematic improvement w.r.t. Sem@$10$. Gains are also reported w.r.t. rank-based metrics in the vast majority of cases.
The only exception is on Yago14k where semantic losses are sometimes slightly outperformed but still competitive w.r.t. their vanilla counterparts. 
However, the difference in terms of MRR and Hits@$10$ values is negligible and not statistically significant. 
This may come from the reduced number of triples in Yago14k, which results in the additional signature information improving Sem@$K$ but not helping discriminate gold entities from others.
Therefore, incorporating BK about relation domains and ranges into $\mathcal{L}^{S}_{PHL}$ and $\mathcal{L}^{S}_{BCEL}$ is a viable approach that provides consistent gains both in terms of MRR, Hits@$10$ and Sem@$10$.
Regarding the benefits from doing so under $\mathcal{L}^{S}_{PLL}$, we can notice a slight decline in MRR and Hits@$10$ values in some cases. However, the other side of the coin is that achieved Sem@$10$ values are substantially higher: except for SimplE on DBpedia77k, Sem@$10$ values increase in a range from $+17\%$ to $+117\%$ for the remaining one-to-one comparisons. As such, our approach using $\mathcal{L}^{S}_{PLL}$ also provides satisfactory results, as long as a slight drop in rank-based metrics is acceptable if it comes with the benefit of significantly better KGEM semantic correctness. Besides, it is worth noting the following points: our hyperparameter tuning strategy relied on the choice of the best $\epsilon$ value on Yago14k -- for computational limitations. The $\epsilon$ value which was found to perform the best on Yago14k was subsequently used in all the remaining scenarios. A more thorough tuning of $\epsilon$ on the other datasets would have potentially provided even more satisfying results under the $\mathcal{L}^{S}_{PLL}$, thus strengthening the value of our approach.

\subsection{Ablation Study}
In this section, KGEMs are tested on three buckets of relations that feature narrow (B1), intermediate (B2), and large (B3) sets of semantically valid heads or tails, respectively. The cut-offs have been manually defined and are provided in supplementary materials\textsuperscript{\ref{git}}. The analysis of B1 allows us to gauge the impact of signature-driven loss functions on relations for which it is harder to predict semantically valid entities. Results reported in Table~\ref{tab:bucket1} for B1 clearly demonstrate that the impact of injecting BK into loss functions is exacerbated for them, thus supporting the value of our approach in a sparse and difficult setting. One might think that the better MRR values achieved with $\mathcal{L}^{S}_{PHL}$, $\mathcal{L}^{S}_{BCEL}$, and $\mathcal{L}^{S}_{PLL}$ are highly correlated to the better Sem@$10$ values. This is partially true, as for a relation in B1, placing all semantically valid candidates at the top of the ranking list is likely to uplift the rank of the ground-truth itself. However, we can see that RGCN-V and RGCN-S have almost equal MRR values on Yago14k, while Sem@$10$ values of RGCN-S are much higher than RGCN-V ($+254\%$). Similar findings hold for ComplEx on Yago14k, ConvE on DBpedia77k, TransE on DBpedia77k and Yago14k. This shows that in a number of cases, signature-driven loss functions improve the semantic correctness of KGEM for small relations while leaving its performance untouched in terms of rank-based metrics. 
Results on B2 and B3 are provided in supplementary materials\textsuperscript{\ref{git}}. In particular, it can be noted that the relative benefit of our approach w.r.t. MRR and Hits@$10$ is more limited on such buckets. Regarding Sem@$K$, results achieved with the vanilla loss functions are already high, hence the relatively lower gain brought by injecting ontological BK. 
These already high Sem@$K$ values may be explained by a higher number of semantically valid candidates. 

\begin{table}
        \caption{Rank-based and semantic-based results on the bucket of relations that feature a narrow set of semantically valid heads or tails (B1). The cut-offs have been manually defined and are provided in supplementary materials.
        }
	\label{tab:bucket1}
    \small
	\centering
                \setlength{\tabcolsep}{0.15cm}
                \begin{adjustbox}{width=0.9\textwidth}
			\begin{tabular}{lccccccccc}
                    \toprule
				&\multicolumn{3}{c}{FB15k187} & \multicolumn{3}{c}{DBpedia77k} & \multicolumn{3}{c}{Yago14k} 
                \\
                    & MRR & H@10 & S@10 &
                    MRR & H@10 & S@10 &
                    MRR & H@10 & S@10 \\
				\toprule
				TransE-V &
    $.535$&$.727$&$.647$& $.498$&$.578$&$.460$& $.914$&$\mathbf{.979}$&$.620$\\
                TransE-S &
    $\mathbf{.646}$&$\mathbf{.805}$&$\mathbf{.937}$& 
    $\mathbf{.539}$&$\mathbf{.626}$&$\mathbf{.910}$& 
    $\mathbf{.922}$&$.975$&$\mathbf{.924}$ \\
    \midrule
                TransH-V &
    $.541$&$.734$&$.661$& $.475$&$.555$&$.425$& $.897$&$.972$&$.436$ \\
                TransH-S &
    $\mathbf{.655}$&$\mathbf{.814}$&$\mathbf{.936}$& 
    $\mathbf{.541}$&$\mathbf{.643}$&$\mathbf{.828}$& 
    $\mathbf{.923}$&$\mathbf{.981}$&$\mathbf{.684}$ \\
    \midrule
                DistMult-V &
    $.589$&$.735$&$.628$& $.466$&$.462$&$.244$& $.949$&$\mathbf{.959}$&$.304$ \\
			  DistMult-S &
    $\mathbf{.667}$&$\mathbf{.802}$&$\mathbf{.929}$& 
    $\mathbf{.498}$&$\mathbf{.547}$&$\mathbf{.451}$& 
    $\mathbf{.965}$&$.958$&$\mathbf{.372}$ \\
    \midrule
    		  ComplEx-V &
    $.530$&$.567$&$.116$& $\mathbf{.424}$&$\mathbf{.425}$&$.161$& $.955$&$.956$&$.133$ \\
    			ComplEx-S &
    $\mathbf{.637}$&$\mathbf{.723}$&$\mathbf{.537}$& 
    $.421$&$.399$&$\mathbf{.198}$& 
    $\mathbf{.961}$&$.956$&$\mathbf{.423}$ \\
    \midrule
                SimplE-V &
    $.507$&$.553$&$.136$& $\mathbf{.396}$&$\mathbf{.370}$&$\mathbf{.273}$& $\mathbf{.959}$&$.882$&$\mathbf{.932}$ \\
                SimplE-S &
    $\mathbf{.576}$&$\mathbf{.671}$&$\mathbf{.505}$& 
    $.324$&$.259$&$.206$& 
    $.958$&$\mathbf{.883}$&$.930$ \\
    \midrule
				ConvE-V &
    $.549$&$.779$&$.973$& $.518$&$\mathbf{.569}$&$.789$& $.969$&$\mathbf{.972}$&$.915$ \\
				ConvE-S &
    $\mathbf{.562}$&$\mathbf{.783}$&$\mathbf{.986}$& 
    $.518$&$.566$&$\mathbf{.927}$& 
    $.969$&$.965$&$\mathbf{.960}$  \\
    \midrule
    			TuckER-V &
    $.597$&$.811$&$.969$& $.519$&$.568$&$.740$& $.949$&$.970$&$.846$ \\
                TuckER-S &
    $\mathbf{.598}$&$\mathbf{.815}$&$\mathbf{.973}$& 
    $\mathbf{.526}$&$\mathbf{.582}$&$\mathbf{.797}$& 
    $\mathbf{.964}$&$.970$&$\mathbf{.892}$ \\
    \midrule
                RGCN-V &
    $.510$&$.629$&$.468$& $.386$&$.387$&$.254$& $.963$&$.959$&$.141$ \\
                RGCN-S &
    $\mathbf{.549}$&$\mathbf{.705}$&$\mathbf{.682}$& 
    $\mathbf{.396}$&$\mathbf{.415}$&$\mathbf{.398}$& 
    $\mathbf{.966}$&$\mathbf{.967}$&$\mathbf{.499}$ \\
    \bottomrule
			\end{tabular}
   \end{adjustbox}
 \end{table}

\section{Discussion}\label{discussion}

\subsection{Treating Different Negatives Differently (RQ1)}

The proposed loss functions $\mathcal{L}^{S}_{PHL}$, $\mathcal{L}^{S}_{BCEL}$, and $\mathcal{L}^{S}_{PLL}$ provide adequate training objective for KGEMs, as evidenced in Table~\ref{tab:schema-defined-results-compiled}. Most importantly, in Section~\ref{lossfunc} we clearly show how the inclusion of signature information into $\mathcal{L}^{S}_{PHL}$, $\mathcal{L}^{S}_{BCEL}$, and $\mathcal{L}^{S}_{PLL}$ can be brought under one roof thanks to a commonly defined semantic factor. Although this semantic factor operates at different levels depending on the loss function, its common purpose is to differentiate how semantically valid and semantically invalid negative triples should be considered compared to positive triples, whereas traditional approaches treat all negative triples indifferently.
Besides, our approach that transforms vanilla loss functions into signature-driven ones can also work for other loss functions such as the pointwise hinge loss or the pairwise logistic loss, as presented in~\cite{mohamed2019}, by including a semantic factor $\epsilon$ as well. 
The tailoring of these losses and the experiments are left for future work.

Recall that compared to complex negative sampler such as KBGAN~\cite{kbgan}, NSCaching~\cite{nscaching}, and self-adversarial NS~\cite{rotate}, we do not introduce any potential overhead due to the need for maintaining a cache~\cite{nscaching} or training an intermediate adversarial learning framework~\cite{kbgan,rotate} for generating high-quality negatives. Instead, negative triples dynamically enter a different part of the loss function depending on their semantic validity. In future work, we will compare the performance and algorithmic complexity of our approach w.r.t. NS procedures, thereby highlighting the potential cost savings in computational resources and execution time compared to sophisticated NS. 
Our approach is also agnostic to the underlying NS procedure, and can work along with simple uniform random NS~\cite{transe} as well as more complex procedures~\cite{kbgan,nscaching,rotate}.
Besides, our approach can be applied even in the absence of BK. In this case, signature-driven loss functions reduce to their vanilla version.

\subsection{Impact of Signature-Driven Losses on Performance (RQ2)}

Mohamed \textit{et al.}~\cite{mohamed2021} investigate the effects of specific choices of loss functions on the scalability and performance of KGEMs w.r.t. rank-based metrics. In this present work, we also assess the semantic capabilities of such models. Based on the results provided and analyzed in Section~\ref{lossfunc}, incorporating BK about relation domains and ranges into the loss functions clearly contribute to a better KGEM semantic correctness, as evidenced by the huge increase frequently observed w.r.t. Sem@$K$. Although it seems intuitive that our proposed approach will result in better predicted semantics, it should not be taken for granted. Hubert \textit{et al.} demonstrate that in the special case of LP on a single target relation (\textit{e.g.}, recommender systems), incorporating schema-based information during training (in their case, during negative sampling) actually decreases the semantic correctness of KGEMs~\cite{ekaw}. This result leads us to posit that injecting ontological information during training does not necessarily lead to a better semantic correctness. The improvement might depend on (1) where this information is consumed (\textit{e.g.}, in the loss function, during negative sampling) and (2) what the task at hand is.

It should also be noted that this increase is not homogeneously distributed across relations: relations with a smaller set of semantically valid entities as heads or tails are more challenging w.r.t. Sem@$K$ (see Table~\ref{tab:bucket1} and results on B2 and B3 in supplementary materials\textsuperscript{\ref{git}}). For such relations, the relative gain from signature-driven loss functions is more acute.
In addition, signature-driven loss functions also drive most of the KGEMs towards better MRR and Hits@$K$ values. As shown in Table~\ref{tab:schema-defined-results-compiled}, this is particularly the case when using the $\mathcal{L}^{S}_{PHL}$ and $\mathcal{L}^{S}_{BCEL}$. This result is particularly interesting, as it suggests that when semantic information about entities and relations is available, there are benefits in using it, even if the intended goal remains to enhance KGEM performance w.r.t. rank-based metrics only.

In addition, it has been noted that including signature information in loss functions has the highest impact on small relations (B1) (Table~\ref{tab:bucket1}), both in terms of rank-based metrics (MRR, Hits@$10$) and semantic correctness (Sem@$10$). For relations with a larger pool of semantically valid entities, the impact is still positive w.r.t. Sem@$10$, but sometimes at the expense of a small drop in terms of rank-based metrics. If the latter metrics are the sole optimization objective, it would be reasonable to design an adaptive training strategy in which vanilla and signature-driven loss functions are alternatively used depending on the current relation and the number of semantically valid candidates as head or tail.

\subsection{On the Evaluation of KGEM Predictions for LP}
\label{section:kgem-eval}

It should be noted that Sem@$K$ measures the capability of models to predict semantically correct triples w.r.t. relation signatures and but does not measure falsehood, contrary to MRR or Hits@$K$.
It can be argued that in specific use-cases, such as e-commerce recommender systems, models should predict the expected item (high MRR/Hits@$K$) but also predict semantically valid items (high Sem@$K$) for a good user experience. 
In other applications such as healthcare, aerospace, or security, users expect to notice when models are failing in order to be able to take over. This corresponds to high MRR/Hits@$K$ and low Sem@$K$ so that errors are clearly noticeable. 

Our experiments and these use cases illustrate the need of both Hits@$K$ and Sem@$K$ metrics to precisely qualify model performance w.r.t. the needs of the considered application (\textit{e.g.}, e-commerce, healthcare).
This opens perspectives on sampling and differently considering different kinds of negative triples to tailor specific aspects of model performance to the requirements of the application.
For instance, based on the results of our experiments, we could envision differently considering negatives for an e-commerce application to obtain a high Sem@$K$ while an aerospace application would require the contrary. 

\section{Conclusion}\label{conclusion}
In this work, we focus on the main loss functions used for link prediction in knowledge graphs. 
Building on the assumption that negative triples are not all equally good for learning better embeddings, we propose to differentiate them based on their semantic validity w.r.t. the domain and range of relations by including relation signature information into loss functions. 
A wide range of KGEMs are subsequently trained under both the vanilla and signature-driven loss functions. In our experiments on three public KGs with different characteristics, the proposed signature-driven loss functions lead to promising results: in most cases, they do not only lead to better MRR and Hits@$10$ values, but also drive KGEMs towards better semantic correctness as measured with Sem@$10$. This  advocates for the further injection of semantic information into loss functions whenever such information is available.
In future work, we will study how the proposed loss functions can accommodate other types of ontological constraints as well as literal nodes.

\bibliographystyle{splncs04}
\bibliography{bibliography}

\clearpage
\appendix\label{appendix}

\section{Modified Versions of $\mathcal{L}^{S}_{BCEL}$ and $\mathcal{L}^{S}_{PLL}$}
\label{modified-versions}
In Table~\ref{tab:epsilon-alpha}, we report results achieved with KGEMs trained under $\mathcal{L}^{S'}_{BCEL}$ and $\mathcal{L}^{S'}_{PLL}$, where the superscript $S'$ denotes a different way to include semantic information. In fact, $\mathcal{L}^{S'}_{PLL}$ makes use of the modified labelling function as used in $\mathcal{L}^{S}_{BCEL}$ and defined in Equation (5) of the paper. Conversely, $\mathcal{L}^{S'}_{BCEL}$ uses the binary (unmodified) labelling function $\ell$ but adopts the same procedure as $\mathcal{L}^{S}_{PLL}$: semantically valid negative triples are considered as positive with probability $\epsilon~\%$.
Hyperparameters for $\mathcal{L}^{S}_{BCEL}$, $\mathcal{L}^{S}_{PLL}$, $\mathcal{L}^{S'}_{BCEL}$, and $\mathcal{L}^{S'}_{PLL}$ are reported in \url{https://github.com/nicolas-hbt/semantic-lossfunc/}.
Results achieved with all the aforementioned loss functions are provided in Table~\ref{tab:epsilon-alpha}. It shows that the signature-driven loss functions presented in the paper are the best performing ones.

\begin{table}[h]
        \caption{Rank-based and semantic-based results on FB15k187, DBpedia77k, and Yago14k. Bold fonts indicate which model performs best w.r.t. a given metric. Suffixes S and S' indicate whether the model is trained under the best (as presented in the paper) or the worst (as presented here) signature-driven version of the loss function, respectively.}
	\label{tab:epsilon-alpha}
	\small
        \centering
                \setlength{\tabcolsep}{0.15cm}
			\begin{tabular}{lccccccccc}
                \toprule
				&\multicolumn{3}{c}{FB15k187} & \multicolumn{3}{c}{DBpedia77k} & \multicolumn{3}{c}{Yago14k} 
                \\
                    & MRR & H@10 & S@10 &
                    MRR & H@10 & S@10 &
                    MRR & H@10 & S@10 \\
				\toprule
    		  ComplEx-S &
    $\mathbf{.316}$&$\mathbf{.476}$&$\mathbf{.796}$& $\mathbf{.297}$&$\mathbf{.409}$&$\mathbf{.897}$& $\mathbf{.923}$&$\mathbf{.931}$&$\mathbf{.667}$ \\
    ComplEx-S' &
    $.227$&$.384$&$.777$& $.252$&$.350$&$.918$& $.907$&$.930$&$.603$ \\
    \midrule
                SimplE-S &
    $\mathbf{.268}$&$\mathbf{.409}$&$.759$& $.230$&$\mathbf{.302}$&$\mathbf{.850}$& $\mathbf{.924}$&$\mathbf{.927}$&$\mathbf{.769}$ \\
    SimplE-S' &
    $.169$&$.288$&$\mathbf{.827}$& $.230$&$.297$&$.583$& $.885$&$.915$&$.290$ \\
    \midrule
			ConvE-S &
    $\mathbf{.283}$&$\mathbf{.476}$&$\mathbf{.996}$& $\mathbf{.283}$&$\mathbf{.405}$&$\mathbf{.985}$& $.933$&$.940$&$\mathbf{.997}$  \\
    ConvE-S' &
    $.271$&$.472$&$.975$& $.273$&$.383$&$.935$& $.933$&$\mathbf{.941}$&$.894$  \\
    \midrule
    		TuckER-S &
    $\mathbf{.320}$&$\mathbf{.522}$&$\mathbf{.996}$& $\mathbf{.312}$&$\mathbf{.421}$&$\mathbf{.969}$& $\mathbf{.931}$&$\mathbf{.943}$&$\mathbf{.929}$ \\
    TuckER-S' &
    $.316$&$.517$&$.983$& $.311$&$.412$&$.912$& $.918$&$.938$&$.867$ \\
    \midrule
                RGCN-S &
    $\mathbf{.260}$&$\mathbf{.415}$&$\mathbf{.860}$& $\mathbf{.197}$&$\mathbf{.320}$&$\mathbf{.957}$& $\mathbf{.927}$&$\mathbf{.934}$&$\mathbf{.828}$ \\
    RGCN-S' &
    $.243$&$.391$&$.780$& $.146$&$.246$&$.862$& $.912$&$.922$&$.385$ \\
    \bottomrule
			\end{tabular}
 \end{table}

\section{Bucket Analysis}\label{appendix:bucket}
Relations are separated into three non-intersecting buckets : relations that feature narrow (B1), intermediate (B2), and large (B3) sets of semantically valid heads or tails, respectively. 
Cut-offs are manually defined for placing a given relation in its corresponding bucket. Such buckets are reported in \url{https://github.com/nicolas-hbt/semantic-lossfunc/}.
Results achieved on B1 are reported in the paper, while results for buckets B2 and B3 for DBpedia77k, FB15k187, and Yago14k are reported in \url{https://github.com/nicolas-hbt/semantic-lossfunc/}.

\end{document}